# A Winner-Take-All Approach to Emotional Neural Networks with Universal Approximation Property

E. Lotfi[1,2], M.-R. Akbarzadeh-T.[2]

[1]*Department of Computer Engineering, Torbat-e-Jam Branch, Islamic Azad University, Torbat-e-Jam, Iran (elotfi@bitools.ir, esilotf@gmail.com)*

[2]*Departments of Electrical and Computer Engineering, Center of Excellence on Soft Computing and Intelligent Information Processing, Ferdowsi University of Mashhad, Iran (akbazar@um.ac.ir)*

**Abstract**— Here, we propose a brain-inspired winner-take-all emotional neural network (WTAENN) and prove the universal approximation property for the novel architecture. WTAENN is a single layered feedforward neural network that benefits from the excitatory, inhibitory, and expandatory neural connections as well as the winner-take-all (WTA) competitions in the human brain's nervous system. The WTA competition increases the information capacity of the model without adding hidden neurons. The universal approximation capability of the proposed architecture is illustrated on two example functions, trained by a genetic algorithm, and then applied to several competing recent and benchmark problems such as in curve fitting, pattern recognition, classification and prediction. In particular, it is tested on twelve UCI classification datasets, a facial recognition problem, three real world prediction problems (2 chaotic time series of geomagnetic activity indices and wind farm power generation data), two synthetic case studies with constant and nonconstant noise variance as well as $k$-selector and linear programming problems. Results indicate the general applicability and often superiority of the approach in terms of higher accuracy and lower model complexity, especially where low computational complexity is imperative.

**References**

[1] Abdi,J., Moshiri, B., Abdulhai, B., Sedigh, A.K., (2011). Forecasting of short-term traffic flow based on improved neuro-fuzzy models via emotional temporal difference learning algorithm. Engineering Applications of Artificial Intelligence, doi:10.1016/j.engappai.2011.09.011

[2] Amin, M. F., Savitha, R., Amin, M. I., & Murase, K. (2012). Orthogonal least squares based complex-valued functional link network. Neural Networks, 32, 257-266.

[3] Babaie, T., Karimizandi, R., & Lucas, C. (2008). Learning based brain emotional intelligence as a new aspect for development of an alarm system. Soft Computing, 12(9), 857-873.

[4] Balkenius, C., Morén J., (2001). Emotional Learning: A Computational Model of AMYG, Cybernetics and Systems, 32(6), 611-636.

[5] Ban, J. C. (2014). Neural network equations and symbolic dynamics. International Journal of Machine Learning and Cybernetics, 1-13.

[6] Binas, J., Rutishauser, U., Indiveri, G., & Pfeiffer, M. (2014). Learning and stabilization of winner-take-all dynamics through interacting excitatory and inhibitory plasticity. Frontiers in computational neuroscience, 8.

[7] Carpenter, G. A., & Grossberg, S. (1987). A massively parallel architecture for a self-organizing neural pattern recognition machine. Computer vision, graphics, and image processing, 37(1), 54-115.

[8] Chandra M., (2005). Analytical Study of A Control Algorithm Based on Emotional Processing, M.S. Dissertation, Indian Institute of Technology Kanpur.

[9] Coultrip, R., Granger, R., & Lynch, G. (1992). A cortical model of winner-take-all competition via lateral inhibition. Neural networks, 5(1), 47-54.

[10] Costarelli, D., & Spigler, R. (2013). Approximation by series of sigmoidal functions with applications to neural networks. Annali di Matematica Pura ed Applicata (1923-), 194(1), 289-306.

[11] Costarelli, D., & Spigler, R. (2013). Approximation results for neural network operators activated by sigmoidal functions. Neural Networks, 44, 101-106.

[12] Costarelli, D. (2014). Interpolation by neural network operators activated by ramp functions. Journal of Mathematical Analysis and Applications, 419(1), 574-582.

[13] Costarelli, D., & Spigler, R. (2013). Multivariate neural network operators with sigmoidal activation functions. Neural Networks, 48, 72-77.

[14] Costarelli, D., & Spigler, R. (2014). Convergence of a family of neural network operators of the Kantorovich type. Journal of Approximation Theory, 185, 80-90.

[15] Costarelli, D. (2015). Neural network operators: Constructive interpolation of multivariate functions. Neural Networks, 67, 28-36.

[16] Coultrip, R., Granger, R., & Lynch, G. (1992). A cortical model of winner-take-all competition via lateral inhibition. Neural networks, 5(1), 47-54.

[17] Cao, H., Cao, F., & Wang, D. (2015). Quantum artificial neural networks with applications. *Information Sciences*, 290, 1-6.

[18] Chen, Y. W., Yang, J. B., Xu, D. L., & Yang, S. L. (2013). On the inference and approximation properties of belief rule based systems. Information Sciences, 234, 121-135.

[19] Cybenko, G. (1989). Approximation by superpositions of a sigmoidal function. Mathematics of control, signals and systems, 2(4), 303-314.

[20] Daryabeigi, E.; Markadeh, G.R.A.; Lucas, C., (2010). Emotional controller (BELBIC) for electric drives — A review, 7-10 Nov., Glendale, AZ, pp: 2901 – 2907, doi: 10.1109/IECON.2010.5674934

[21] Dalgleish, Tim. (2004). The emotional brain. Nature Reviews Neuroscience, vol. 5, no. 7, pp. 583-589.

[22] Dehkordi, B. M., Parsapoor, A., Moallem, M., & Lucas, C. (2011a). Sensorless speed control of switched reluctance motor using brain emotional learning based intelligent controller. Energy Conversion and Management, 52(1), 85-96.

[23] Dehkordi, B. M., Kiyoumarsi, A., Hamedani, P., & Lucas, C. (2011b). A comparative study of various intelligent based controllers for speed control of IPMSM drives in the field-weakening region. Expert Systems with Applications, 38(10), 12643-12653.

[24] Dehuri, S., & Cho, S. B. (2010). Evolutionarily optimized features in functional link neural network for classification. Expert Systems with Applications, 37(6), 4379-4391.

[25] Dolcos, F., & McCarthy, G. (2006). Brain systems mediating cognitive interference by emotional distraction. The Journal of Neuroscience, 26(7), 2072-2079.

[26] Dorrah, H. T., El-Garhy, A. M., & El-Shimy, M. E. (2011). PSO-BELBIC scheme for two-coupled distillation column process. Journal of advanced Research, 2(1), 73-83.

[27] Dehuri, S., Roy, R., Cho, S. B., & Ghosh, A. (2012). An improved swarm optimized functional link artificial neural network (ISO-FLANN) for classification. Journal of Systems and Software, 85(6), 1333-1345.

[28] Douglas, Rodney J., and Kevan AC Martin. (2004), Neuronal circuits of the neocortex. Annu. Rev. Neurosci. 27: 419-451.

[29] [29]Fino, E., & Yuste, R. (2011). Dense inhibitory connectivity in neocortex. Neuron, 69(6), 1188-1203.

[30] Gripenberg, G. (2003). Approximation by neural networks with a bounded number of nodes at each level. Journal of approximation theory, 122(2), 260-266.

[31] Grossberg S., Seidman, D. (2006) Neural dynamics of autistic behaviors: Cognitive, emotional, and timing substrates. Psychological Review, 113,483–525.

[32] Hall, John E. Guyton and Hall Textbook of Medical Physiology: Enhanced E-book. Elsevier Health Sciences, 2010.

[33] Haykin, S., & Netwo, N. (2004). A comprehensive foundation. Neural Networks, 2(2004).

[34] Horton, C. W., & Ichikawa, Y. H. (1996). Chaos and structures in nonlinear plasmas (Vol. 238). Singapore: World Scientific

[35] Hornik, K., Stinchcombe, M., & White, H. (1989). Multilayer feedforward networks are universal approximators. Neural networks, 2(5), 359-366.

[36] Horton, W. (1997). Chaos and structures in the magnetosphere. Physics reports, 283(1), 265-302

[37] Horton, W., Smith, J. P., Weigel, R., Crabtree, C., Doxas, I., Goode, B., & Cary, J. (1999). The solar-wind driven magnetosphere–ionosphere as a complex dynamical system. Physics of Plasmas (1994-present), 6(11), 4178-4184.

[38] Huang, H., & Wu, C. (2009). Approximation capabilities of multilayer fuzzy neural networks on the set of fuzzy-valued functions. Information Sciences, 179(16), 2762-2773.

[39] Hassoun, M. (1995) Fundamentals of Artificial Neural Networks MIT Press, p. 48

[40] Hornik, K. (1991). Approximation capabilities of multilayer feedforward networks. Neural networks, 4(2), 251-257.

[41]     Hashem, S. (1997). Optimal linear combinations of neural networks. Neural networks, 10(4), 599-614.
[42]     Hussein Al-Arashi, W., Ibrahim, H., & Azmin Suandi, S. (2014). Optimizing principal component analysis performance for face recognition using genetic algorithm. Neurocomputing. 128, 415-420.
[43]     Ismailov, V. E. (2014). On the approximation by neural networks with bounded number of neurons in hidden layers. Journal of Mathematical Analysis and Applications, 417(2), 963-969.
[44]     Itti, L., Koch, C., & Niebur, E. (1998). A model of saliency-based visual attention for rapid scene analysis. IEEE Transactions on pattern analysis and machine intelligence, 20(11), 1254-1259.
[45]     Jafarzadeh, S., (2008). Designing PID and BELBIC Controllers in Path Tracking Problem, Int. J. of Computers, Communications & Control, ISSN 1841-9836, E-ISSN 1841-9844, Vol. III (2008), Suppl. issue: Proceedings of ICCCC 2008, pp. 343-348.
[46]     Jothi, R. G., & Rani, S. M. (2014). Hybrid neural network for classification of graph structured data. International Journal of Machine Learning and Cybernetics, 6(3), 465-474.
[47]     Kainen, P. C., & Kurková, V. (2009). An integral upper bound for neural network approximation. Neural Computation, 21(10), 2970-2989.
[48]     Khalghani, M. R., & Khooban, M. H. (2014). A novel self-tuning control method based on regulated bi-objective emotional learning controller's structure with TLBO algorithm to control DVR compensator. Applied Soft Computing, 24, 912-922.
[49]     Khashman, A., (2008). A modified back propagation learning algorithm with added emotional coefficients, IEEE Trans. Neural Networks, 19(11), 1896-1909.
[50]     Khashman, A. (2010). Modeling cognitive and emotional processes: A novel neural network architecture. Neural Networks, 23(10), 1155-1163.
[51]     Khashman, A. (2012). An emotional system with application to blood cell type identification. Transactions of the Institute of Measurement and Control, 34(2-3), 125-147.
[52]     Kalayci, T. E., Bahrepour, M., Meratnia, N., & Havinga, P. J. (2011). How Wireless Sensor Networks Can Benefit from Brain Emotional Learning Based Intelligent Controller (BELBIC). Procedia Computer Science, 5, 216-223.
[53]     Khalilian, M., Abedi, A., & Zadeh, A. D. (2012). Position control of hybrid stepper motor using brain emotional controller. Energy Procedia, 14, 1998-2004.
[54]     Kohonen, Teuvo. (1990). The self-organizing map. Proceedings of the IEEE 78, no. 9 (1990): 1464-1480.
[55]     Khosravi, A., Nahavandi, S., Creighton, D., & Atiya, A. F. (2011). Lower upper bound estimation method for construction of neural network-based prediction intervals. Neural Networks, IEEE Transactions on, 22(3), 337-346.
[56]     LeDoux, J., (1996). The emotional brain. Simon and Schuster, New York.
[57]     LeDoux, J. E., (2000). Emotion circuits in the brain. Annual review of neuroscience, 23(1), 155-184.
[58]     Li, S., Liu, B., & Li, Y. (2013). Selective Positive–Negative Feedback Produces the Winner-Take-All Competition in Recurrent Neural Networks. Neural Networks and Learning Systems, IEEE Transactions on, 24(2), 301-309.
[59]     Liu, Q., & Wang, J. (2008). Two k-winners-take-all networks with discontinuous activation functions. Neural Networks, 21(2), 406-413.

[60] Liu, Q., Dang, C., & Cao, J. (2010). A Novel Recurrent Neural Network With One Neuron and Finite-Time Convergence for-Winners-Take-All Operation. Neural Networks, IEEE Transactions on, 21(7), 1140-1148.

[61] Liu, Q., Cao, J., & Chen, G. (2010). A novel recurrent neural network with finite-time convergence for linear programming. Neural computation, 22(11), 2962-2978.

[62] Lin, B. S., Lin, B. S., Chong, F. C., & Lai, F. (2006). A functional link network with higher order statistics for signal enhancement. Signal Processing, IEEE Transactions on, 54(12), 4821-4826.

[63] Lotfi, E., & Akbarzadeh-T, M. R. (2014b). Adaptive brain emotional decayed learning for online prediction of geomagnetic activity indices. Neurocomputing, 126, 188-196.

[64] Lotfi, E., & Keshavarz, A. (2014). Gene expression microarray classification using PCA–BEL. Computers in biology and medicine, 54, 180-187.

[65] Lotfi, E., M. R. Akbarzadeh-T., (2014a). Practical emotional neural networks. Neural Networks. doi: 10.1016/j.neunet.2014.06.012

[66] Lotfi, E., Setayeshi, S., & Taimory, S. (2014). A neural basis computational model of emotional brain for online visual object recognition. Applied Artificial Intelligence, 28(8), 814-834.

[67] Lucas, C., Shahmirzadi, D., and Sheikholeslami, N., (2004). Introducing BELBIC: Brain emotional learning based intelligent controller. International Journal of Intelligent Automation and Soft Computing, 10, 11-21.

[68] Lucas, C. (2011). BELBIC and its industrial applications: towards embedded neuroemotional control codesign. In Integrated Systems, Design and Technology 2010 (pp. 203-214). Springer Berlin Heidelberg.

[69] Ma, L., & Khorasani, K. (2004). New training strategies for constructive neural networks with application to regression problems. Neural Networks, 17(4), 589-609.

[70] Malsburg von der, Chr. (1973). Self-organization of orientation sensitive cells in the striate cortex. Kybernetik 14, no. 2 (1973): 85-100.

[71] Makovoz, Y. (1998). Uniform approximation by neural networks. Journal of Approximation Theory, 95(2), 215-228.

[72] Marinov, C., & Hopfield, J. J. (2005). Stable computational dynamics for a class of circuits with O (N) interconnections capable of KWTA and rank extractions. Circuits and Systems I: Regular Papers, IEEE Transactions on, 52(5), 949-959.

[73] Mehrabian, A. R., Lucas, C., & Roshanian, J. (2006). Aerospace launch vehicle control: an intelligent adaptive approach. Aerospace Science and technology, 10(2), 149-155.

[74] Mehrabian, A. R. and Lucas, C., (2005). Emotional Learning based Intelligent Robust Adaptive Controller for Stable Uncertain Nonlinear Systems, International Journal of Engineering and Mathematical Sciences, 2(4), 246-252.

[75] Misra, B. B., & Dehuri, S. (2007). Functional link artificial neural network for classification task in data mining. Journal of Computer Science, 3(12), 948.

[76] Morén, J. and C. Balkenius, (2000). A Computational Model of Emotional Learning in the AMYG. In: From Animals to Animats 6: Proceedings of the 6th International Conference on the Simulation of Adaptive Behaviour, Meyer, J.A., A. Berthoz, D. Floreano, H.L. Roitblat and S.W. Wilson (Eds.). MIT Press, Cambridge, MA., USA., pp: 115-124.

[77] Morén, J., (2002). Emotion and learning-A computational model of the AMYG. Ph.D. Thesis, Department of Cognitive Science, Lund University, Lund, Sweden.

[78] Palm, G., & Sommer, F. T. (1992). Information capacity in recurrent McCulloch-Pitts networks with sparsely coded memory states. Network: Computation in Neural Systems, 3(2), 177-186.

[79] Papadopoulos, G., Edwards, P. J., & Murray, A. F. (2001). Confidence estimation methods for neural networks: A practical comparison. Neural Networks, IEEE Transactions on, 12(6), 1278-1287.

[80] Phelps, E. A. (2006). Emotion and cognition: insights from studies of the human amygdala. Annu. Rev. Psychol., 57, 27-53.

[81] Pavlos, G. P., Iliopoulos, A. C., Tsoutsouras, V. G., Sarafopoulos, D. V., Sfiris, D. S., Karakatsanis, L. P., & Pavlos, E. G. (2011). First and second order non-equilibrium phase transition and evidence for non-extensive Tsallis statistics in Earth's magnetosphere. Physica A: Statistical Mechanics and its Applications, 390(15), 2819-2839.

[82] Quan, Hao, DiptiSrinivasan, and Abbas Khosravi, (2014). Short-Term Load and Wind Power Forecasting Using Neural Network-Based Prediction Intervals. Neural Networks and Learning Systems, IEEE Transactions on, 99: 2013: 1-1.

[83] Riesenhuber, Maximilian, and Tomaso Poggio (1999). Hierarchical models of object recognition in cortex. Nature neuroscience 2(11): 1019-1025.

[84] Rouhani, H., Jalili, M., Araabi, B. N., Eppler, W., & Lucas, C. (2007). Brain emotional learning based intelligent controller applied to neurofuzzy model of micro-heat exchanger. Expert Systems with Applications, 32(3), 911-918.

[85] Reifman, J., & Feldman, E. E. (2002). Multilayer perceptron for nonlinear programming. Computers & Operations Research, 29(9), 1237-1250. Misra, B. B., & Dehuri, S. (2007). Functional link artificial neural network for classification task in data mining. Journal of Computer Science, 3(12), 948.

[86] Sadeghieh, A., Sazgar, H., Goodarzi, K., & Lucas, C. (2012). Identification and real-time position control of a servo-hydraulic rotary actuator by means of a neurobiologically motivated algorithm. ISA transactions, 51(1), 208-219.

[87] Tamura, S. I., & Tateishi, M. (1997). Capabilities of a four-layered feedforward neural network: four layers versus three. Neural Networks, IEEE Transactions on, 8(2), 251-255.

[88] Tsotsos, John K., Scan M. Culhane, Winky Yan Kei Wai, Yuzhong Lai, Neal Davis, and Fernando Nuflo (1995). Modeling visual attention via selective tuning. Artificial intelligence 78 (1): 507-545.

[89] Wu, A., Zeng, Z., & Chen, J. (2014). Analysis and design of winner-take-all behavior based on a novel memristive neural network. Neural Computing and Applications, 24(7-8), 1595-1600

[90] Wu, H., Wang, K., Guo, Q., Xu, G., & Li, N. (2014). Design of a kind of nonlinear neural networks for solving the inverse optimal value problem with convex constraints. International Journal of Machine Learning and Cybernetics, 5(1), 85-92.

[91] Wu, H., Zhang, X., Li, R., & Yao, R. (2015). Adaptive exponential synchronization of delayed Cohen–Grossberg neural networks with discontinuous activations. International Journal of Machine Learning and Cybernetics, 6(2), 253-263.

[92] Yin, F., Jiao, L. C., Shang, F., Xiong, L., & Wang, X. (2014). Sparse regularization discriminant analysis for face recognition. Neurocomputing, 128, 341-362.

[93] Yong, E. M., Qian, W. Q., & He, K. F. (2014). An adaptive predictor–corrector reentry guidance based on self-definition way-points. Aerospace Science and Technology, 39, 211-221.

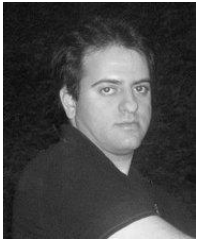
**Ehsan Lotfi** is Assistant Professor of Department of Computer Engineering at Azad University, Torbat-e-Jam branch, Torbat-e-Jam, Iran. He received the B.Sc. degree in Computer Engineering (2006), from Ferdowsi University of Mashhad, M.Sc. degree in Artificial Intelligence (2009) from Azad University, Mashhad Branch and his PhD in Artificial Intelligence from Science and Research campus of Azad University. From 2006-to-2008 he was research assistant in Khorasan Research Center for Advance technology of Iran. His research interest includes cognitive sciences, computational and artificial intelligence, soft computing and their applications. He has published 12 peer-reviewed articles and a national patent in his research programs.

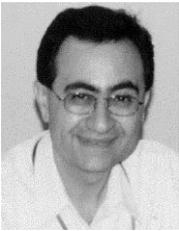
**Mohammad-R. Akbarzadeh-T.** (Senior Member, IEEE) is a professor of electrical engineering and computer engineering at Ferdowsi University of Mashhad. He received his PhD on *Evolutionary Optimization and Fuzzy Control of Complex Systems* from the department of electrical and computer engineering at the University of New Mexico in 1998. In 2006-2007, he also joined BISC-UC Berkeley as a visiting scholar. In 2011, he chaired the first National Workshop on Soft Computing and Intelligent Systems in Mashhad. In 2007, he served as the technical chair for the First Joint Congress on Fuzzy & Intelligent Systems that was held in Mashhad, Iran.

Dr. Akbarzadeh is the founding president of the Intelligent Systems Scientific Society of Iran and the founding councilor representing the Iranian Coalition on Soft Computing in IFSA. From 2000-to-2008, he served as the faculty advisor for the IEEE student branch at Ferdowsi University of Mashhad. He has received several awards including: the IDB Excellent Leadership Award in 2010, the Outstanding Faculty Awards in 2008 and 2002. His research interests include evolutionary algorithms, fuzzy logic and control, soft computing, multi-agent systems, complex systems, robotics, and biomedical engineering systems. He has published over 300 peer-reviewed articles in these and related research fields.